\documentclass{sig-alternate-05-2015}
\usepackage{amsmath}
\usepackage{bm}
\DeclareMathOperator{\Tr}{Tr}
\DeclareMathOperator{\diag}{diag}
\DeclareMathOperator{\rank}{rank}
\DeclareMathOperator{\loss}{loss}

\begin{document}

%
%
%
%
%
%
\CopyrightYear{2016}
\setcopyright{acmcopyright}
\conferenceinfo{SIGIR '16,}{July 17-21, 2016, Pisa, Italy}
\isbn{978-1-4503-4069-4/16/07}\acmPrice{\$15.00}
\doi{http://dx.doi.org/10.1145/2911451.2914693}

\title{Subspace Clustering Based Tag Sharing for Inductive Tag Matrix Refinement with Complex Errors}
\numberofauthors{3}
%
\author{Yuqing Hou$^{\dag}$\qquad Zhouchen Lin\titlenote{Corresponding author.}$^{\dag}$\qquad Jin-ge Yao$^{\S}$\\
$^\dag$\affaddr{Key Lab. of Machine Perception (MOE), School of EECS, Peking University, P. R. China}\\
$^\S$\affaddr{Institute of Computer Science and Technology, Peking University, P. R. China}\\
\email{\{yqh, zlin, yaojinge\}@pku.edu.cn}
}

\maketitle
\begin{abstract}
Annotating images with tags is useful for indexing and retrieving images.
However, many available annotation data include missing or inaccurate annotations.
In this paper, we propose an image annotation framework which sequentially performs tag completion and refinement. We utilize the subspace property of data via sparse subspace clustering for tag completion. Then we propose a novel matrix completion model for tag refinement, integrating visual correlation, semantic correlation and the novelly studied property of complex errors. The proposed method outperforms the state-of-the-art approaches on multiple benchmark datasets even when they contain certain levels of annotation noise.
\end{abstract}

%
%

\vspace{-4mm}
\vspace{-4mm}
\section{Introduction}
\label{sec:intro}
It is useful to annotate images with textual tags for the purpose of image indexing and retrieval. To annotate proper tags, one need to bridge the gap between low level visual features of an image and corresponding high level semantic information 
\cite{li2015socializing}. Since manual annotation is labor intensive, automatic annotation has aroused much attention. 
Many machine learning based approaches have been developed.

Currently many image annotation data have been collected from crowdsourcing services \cite{russell2008labelme,huiskes08}, providing large amount of data for training while being noisy due to annotation errors. Annotation errors are usually complex 
and mainly come in two forms: missing tags and inaccurate tags.
Most image annotation approaches solely focus on one of those two, either trying to impute the missing tags (tag completion/tag assignment)
\cite{guillaumin2009tagprop} or correcting inaccurate tags (tag refinement)
\cite{wu2013tag,feng2014image,li2015socializing}.
Other existing methods fail to model the complex errors properly. They either treat them in the same way \cite{zhu2010image}, ignoring the complex property of the errors, or rigidly assign fixed weights to different kinds of errors \cite{guillaumin2009tagprop}, having no adaptability when working on different datasets with different levels of annotation errors. 

In this paper, we propose a framework called Subspace clustering and Matrix completion with Complex errors (SMC). Since current tag refinement methods suffer from the extreme sparsity problem 
\cite{natarajan2014inductive}, SMC performs tag completion and refinement sequentially.
During tag completion, SMC tries to introduce many additional proper tags to images
via exploring subspace property in the image collection.
We then adapt the inductive matrix completion \cite{jain2013provable} model to perform the following tag refinement procedure, utilizing side information such as the correlation between visual features and their corresponding tags (visual correlation), correlation between the semantic information of tags (semantic correlation) and the complex errors.

The main contributions of this paper include:
\begin{itemize}
    \item We perform tag completion and tag refinement sequentially, showing that tag refinement benefits from tag completion.
	\item	We formulate tag completion in a subspace clustering framework to tackle the extreme sparsity problem. 
	\item	We novelly adapt the inductive matrix completion model for tag refinement, taking visual correlation, semantic correlation and our novelly studied complex errors property into consideration.
\end{itemize}

\vspace{-4mm}
\section{The Proposed Framework}
\subsection{Overview}
We denote the observed tag matrix as $\mathbf{O}\in \{0,1\}^{N_{i}\times N_{t}}$, where each row corresponds to one image, each column corresponds to one textual tag, and $N_{i}$ and $N_{t}$ denote the number of images and tags, respectively. $\mathbf{O}_{ij}$ takes value $1$ only if image $i$ is annotated with tag $j$ and $0$ otherwise.

We are targeting at modifying the values in matrix $\mathbf{O}$ by matrix completion methods to perform image annotation.
After the matrix completion procedure, if 
the value of $\mathbf{O}_{ij}$ changes from nonzero (zero) to zero (nonzero),
we say that the algorithm removes (adds) tag $j$ from (to) image $i$. Methods based on matrix completion are robust and efficient since they only operate on the tag matrix, avoiding error propagation from image segmentation.

However, in many cases $\mathbf{O}$ is so sparse that some columns have at most one known entries and some rows have no known entries at all, making existing methods not applicable \cite{natarajan2014inductive}. In order to overcome such extreme sparsity, we first perform tag completion to make $\mathbf{O}$ denser, creating a better condition for the following tag refinement procedure. More specifically, we perform subspace clustering over images and share tags within subspaces.

For tag refinement, existing methods usually depends heavily on image segmentation and visual feature extraction accuracies \cite{carneiro2007supervised}. However, image segmentation and feature extraction procedures always contain a lot of noises, which affect the following annotation procedure severely. Meanwhile, recent matrix completion based methods \cite{goldberg2010transduction,feng2014image,feng2013large} stand out due to their robustness and efficiency, since these algorithms avoid the image segmentation procedure.

Our proposed framework is called Subspace clustering and Matrix completion with Complex errors (SMC),
because it utilizes the subspace property of image collections (Section~\ref{sec:sc})
and addresses the complex errors and side information in an inductive matrix completion model for tag refinement (Section~\ref{sec:mc}).

\vspace{-1mm}
\subsection{Subspace Clustering for Tag Completion}\label{sec:sc}
\vspace{-1mm}
\subsubsection{Subspace Clustering}
It is 
reasonable to assume that images belonging to different categories are approximately sampled from a mixture of several low-dimensional subspaces. The membership of the data points to the subspaces is unobserved, leading to the challenging problem of subspace clustering. Here the goal is to cluster data into $k$ clusters with each cluster corresponding to a subspace.

One of the state-of-the-art method is the sparse subspace clustering (SSC) model \cite{elhamifar2009sparse}.
The idea behind SSC is to express a data point as a linear (or affine)
combination of neighboring data points. The neighbors can be any other
points in the data set. While every point is a combination of all other data points, SSC seeks for the sparsest representation among all the candidates by minimizing the number of nonzero coefficients \cite{elhamifar2010clustering}.

We denote the set of images, represented as visual feature vectors, as $\mathbf{V} =[\mathbf{v}_1,\mathbf{v}_2,\dots,\mathbf{v}_n]$. Assuming that they are drawn from a union of $k$ subspaces. Each column of $\mathbf{V}$ can be represented by a linear combination of the bases in a ``dictionary". SSC uses the matrix $\mathbf{V}$ itself as the dictionary while explicitly considering noise:
\begin{eqnarray}
 & \min\limits_{\mathbf{Z},\mathbf{E} } & \|\mathbf{Z}\|_{1} + \mu \|\mathbf{E}\|^{2}_{F},\\
 & s.t. & \mathbf{V} = \mathbf{VZ}^\top + \mathbf{E}, \diag(\mathbf{Z}) = \mathbf{0}, \mathbf{Z1} = \mathbf{1},
\end{eqnarray}
where $\mathbf{Z}^\top =[\mathbf{z}_1,\mathbf{z}_2,\dots,\mathbf{z}_n]$ is the coefficient matrix with each $\mathbf{z}_i$ being the representation of $\mathbf{v}_i$ and $\mathbf{E}$ is the error matrix. This problem can be solved efficiently using modern sparse optimization algorithms, such as linearized alternating direction methods \cite{lin2011linearized}.

Given a sparse representation for each data point, we can define the affinity matrix as
$\mathbf{A} = |\mathbf{Z}| + |\mathbf{Z}^\top|$.
Subspaces are then obtained by applying spectral clustering to the Laplacian matrix of $\mathbf{A}$ \cite{elhamifar2009sparse}.
\subsubsection{Tag Sharing}
We improve the 
search based neighbor voting algorithm proposed in \cite{makadia2008new} to share tags in each cluster separately. We rank all the tags for the cluster, taking tag frequency, tag co-occurrence and local frequency into consideration. The elements of tag matrix after tag sharing are no longer binary but take values in $[0,1]$, representing the confidence level between each image-tag pair.

\subsection{Matrix Completion for Tag Refinement}\label{sec:mc}
The tag completion procedure makes the tag matrix much denser and thus avoids the extreme sparsity problem.
Then we can refine the tag matrix.
In our framework we novelly adapt the inductive matrix completion model (IMC) \cite{jain2013provable} for tag refinement,
due to its scalability and capability of incorporating various kinds of side information.
\vspace{-1mm}
\subsubsection{Inductive Matrix Completion}
Let $\mathbf{v}_i\in \mathbb{R}^{f_{i}}$ denote the $f_{i}$-dimensional feature vector of image $i$ and $ \mathbf{t}_j \in \mathbb{R}^{f_{t}}$ denote the $f_{t}$-dimensional feature vector of tag $j$. Let $ \mathbf{V} \in \mathbb{R}^{N_{i}\times f_{i}}$ denote the feature matrix of $N_{i}$ images, where the $i$-th row is the image feature vector $\mathbf{v}_i^\top$, and $\mathbf{T} \in \mathbb{R}^{N_{t}\times f_{t}}$ denote the feature matrix of $N_{t}$ tags, where the $i$-th row is the tag feature $\mathbf{t}_i^\top$.

For image annotation, we assume that the tag matrix can be approximated by applying visual feature vectors and tag feature vectors associated with its row and column entries onto an underlying low-rank matrix $\mathbf{M}$, i.e. $\mathbf{O}\approx \mathbf{VMT}^\top$, where $\mathbf{M} = \mathbf{PQ}^\top$ \cite{jain2013provable} and $\mathbf{P} \in \mathbb{R}^{f_{i} \times r}$ and $\mathbf{Q} \in \mathbb{R}^{ r \times f_{t}}$ are of rank $r \ll N_{i}, N_{t}$.
The goal is to solve the following problem:
\begin{equation}\label{opt:imc}
	\min_{\mathbf{P},\mathbf{Q} }\quad\loss(\mathbf{O}, \mathbf{V}\mathbf{PQ}^\top\mathbf{T}^\top) +
\lambda_1 ( \rank(\mathbf{PQ}^\top) ).
\end{equation}

A common choice for the loss function is the squared loss. The low-rank constraint on $\mathbf{PQ}^\top$ makes (\ref{opt:imc}) NP-hard. A standard relaxation is to use the trace norm, i.e. sum of singular values. Minimizing the trace-norm of $\mathbf{M} = \mathbf{PQ}^\top$ is equivalent to minimizing $\frac{1}{2}(\|\mathbf{P}\|^{2}_{F} + \|\mathbf{Q}\|^{2}_{F})$ \cite{jain2013provable}.
The relaxed optimization problem we use in this work is therefore:
\begin{equation}
\min_{\mathbf{P},\mathbf{Q}}\quad
\|\mathbf{O} - \mathbf{V}\mathbf{PQ}^\top\mathbf{T}^\top\|^{2}_F + \frac{\lambda_1}{2}(\|\mathbf{P}\|^{2}_{F} + \|\mathbf{Q}\|^{2}_{F}).
\end{equation}

\subsubsection{Visual Correlation}
We want to get the refined tag matrix $\mathbf{\hat{O}} = \mathbf{V}\mathbf{PQ}^\top\mathbf{T}^\top $ from the original tag matrix $\mathbf{O}$. Here we represent the $i$th row of $\mathbf{\hat{O}}$ as $\mathbf{\hat{O}}_i$, corresponding to the refined tag vector of image $i$. Thus we can measure the correlation between image $i$ and image $j$ in two ways: $1)$ similarity between image features $\mathbf{v}_i$ and $\mathbf{v}_j$, $2)$ similarity between refined tag vectors $\mathbf{\hat{O}}_i$ and $\mathbf{\hat{O}}_j$. Since visually similar images often belong to similar themes and thus are annotated with similar tags, these two kinds of similarities should be correlated.

Such visual correlation can be enforced by solving the following optimization
\begin{equation}
   \min_{\mathbf{P},\mathbf{Q}}\quad\sum_{i=1}^{N_{t}}\sum_{j=1}^{N_{t}} \| \mathbf{\hat{O}}_i - \mathbf{\hat{O}}_j \|^2g_{ij}  ,
\end{equation}
where $\| \mathbf{\hat{O}}_i - \mathbf{\hat{O}}_j \|^2$ measures the similarity between tag vectors $\mathbf{\hat{O}}_i$ and $\mathbf{\hat{O}}_j$ and $g_{ij}$ measures the similarity between visual features $\mathbf{v}_i$ and $\mathbf{v}_j$. In this work, we adopt cosine similarity, i.e. $g_{ij} = \cos(\mathbf{v}_i, \mathbf{v}_j)$. The formulation forces tag vectors with large similarities also have large similarity in their corresponding visual features and vice versa.

The formulation can be rewritten as
\begin{equation}\label{opt:vcorr}
\min_{\mathbf{P},\mathbf{Q} }\Tr(\mathbf{\hat{O}L_v\hat{O}}^\top) = \min\limits_{\mathbf{P},\mathbf{Q} }\Tr(\mathbf{V}\mathbf{PQ}^\top\mathbf{T}^\top\mathbf{L_{v}T}\mathbf{QP}^\top\mathbf{V}^\top),
\end{equation}
where $\mathbf{L_v} = \diag(\mathbf{G}\mathbf{1}) - \mathbf{G}$ is the Graph Laplacian \cite{chung1997spectral} of the similarity matrix $\mathbf{G}=(g_{ij})$.
\vspace{-1mm}
\subsubsection{Semantic Correlation}
Similarly, we can also enforce semantic correlation between tags.
Since each column of the matrix $\mathbf{\hat{O}}$ represents the feature of a tag, we can measure the correlation between two tags using the similarity between their corresponding column vectors of $\mathbf{\hat{O}}$.
Meanwhile, semantic similarity between two tags can be measured using word vectors.
These two kinds of similarities should be correlated as well.

We can enforce the semantic correlation by solving the following optimization, in a similar form as (\ref{opt:vcorr}):
\begin{eqnarray}
\min_{\mathbf{P},\mathbf{Q} }\Tr(\mathbf{\hat{O}}^\top\mathbf{L_{s}\hat{O}}) = \min_{\mathbf{P},\mathbf{Q}}\Tr(\mathbf{T}\mathbf{QP}^\top\mathbf{V}^\top\mathbf{L_{s}V}\mathbf{PQ}^\top\mathbf{T}^\top),
\end{eqnarray}
where $\mathbf{L_s} = \diag(\mathbf{H}\mathbf{1}) - \mathbf{H}$ is the Graph Laplacian of the similarity matrix $\mathbf{H}=(h_{ij})$, with each element $h_{ij}=\cos(\mathbf{t}_i, \mathbf{t}_j)$.

\vspace{-1mm}
\subsubsection{Features Vectors}
We utilize DeCAF$_{6}$ \cite{donahue2013decaf} to extract $4,096$-dimensional visual features for each image, which have high level 
information. 
Meanwhile, we adopt pre-trained word embedding vectors (word2vec) \cite{mikolov2013efficient} to construct $300$-dimensional features for each tag, trying to capture semantic information.
\vspace{-1mm}
\subsubsection{Complex Errors}
As we have mentioned, annotation errors come in two forms: missing tags and inaccurate tags. Since human beings are relatively reasonable, the user-provided tags are reasonably accurate to certain level \cite{zhu2010image}.
Users might miss one or several proper tags among the few related tags, but may become less probable to add one or several inaccurate tags from the massive unrelated tag sets \cite{huiskes08}.
In other words, if an image is not originally annotated with a tag, it is more likely that they really have no relation at all.
Thus the errors are mainly composed of inaccurate tags rather than missing tags. And we should pay more attention to denoise the inaccurate tags rather than completing the missing ones.

To model the complex structure of errors, we improve the matrix completion model by putting less weights on the unannotated positions:
\begin{eqnarray}
 \min_{\mathbf{P},\mathbf{Q}} \|\mathbf{O} - \mathbf{V}\mathbf{PQ}^\top\mathbf{T}^\top \|^{2}_F
 - \mu \|\mathbf{U_{\Omega}} (\mathbf{O} - \mathbf{V}\mathbf{PQ}^\top\mathbf{T}^\top)\|^{2}_{F},
\end{eqnarray}
where $\mathbf{\Omega}$ represents the positions where the images are originally not annotated. $\mathbf{U}$ is a projection operator and $\mu$ acts as a weighting parameter which changes adaptively in different datasets according to their noise levels.

Existing methods never model these two kinds of errors separately. They simply model the errors as Laplacian noise \cite{zhu2010image} or Gaussian noise \cite{wu2013tag}. To our knowledge, our model is the first to model the missing errors and inaccurate errors separately. The model can further adapt to different datasets according to their noise levels.
\vspace{-4mm}
\section{Final Model}
Based on the components regarding low-rankness, visual correlation, semantic correlation and complex errors, we formulate the objective function as follows:
\begin{gather*}
\min_{\mathbf{P},\mathbf{Q}} \|\mathbf{O} - \mathbf{V}\mathbf{PQ}^\top\mathbf{T}^\top \|^{2}_F
 - \mu\|\mathbf{U_{\Omega}} (\mathbf{O} - \mathbf{V}\mathbf{PQ}^\top\mathbf{T}^\top)\|^{2}_{F} \\
 + \frac{\lambda_1}{2}(\|\mathbf{P}\|^{2}_{F} + \|\mathbf{Q}\|^{2}_{F}) + \\
\lambda_2 [\Tr(\mathbf{V}\mathbf{PQ}^\top\mathbf{T}^\top\mathbf{L_{v}T}\mathbf{QP}^\top\mathbf{V}^\top\!+\! \mathbf{T}\mathbf{QP}^\top\mathbf{V}^\top\mathbf{L_{s}V}\mathbf{PQ}^\top\mathbf{T}^\top)].
\end{gather*}
By solving $\mathbf{P}$ and $\mathbf{Q}$ we can then construct the refined tag matrix $\mathbf{\hat{O}} = \mathbf{V}\mathbf{PQ}^\top\mathbf{T}^\top$ and use it for refined annotation.

We set the regularization terms of visual correlation and semantic correlation with the same weight $\lambda_2$
for simplicity. This simplification does not harm performance, as we find during preliminary experiments.

This objective function is non-convex. To solve the optimization problem, we adapt the solver for low-rank empirical risk minimization for multi-label learning (LEML) \cite{yu2014large}, which naturally fits for the settings of large-scale multi-label learning with missing labels. The solver uses alternating minimization (fix $\mathbf{P}$ and solve for $\mathbf{Q}$ and vice versa) to update the variables. When either $\mathbf{P}$ or $\mathbf{Q}$ is fixed, the resulting subproblem in one variable ($\mathbf{Q}$ or $\mathbf{P}$) can be solved using iterative conjugate gradient procedure.

\section{Experimental Evaluation}
\subsection{Datasets and Experimental Setup}
We evaluate our proposed SMC framework on two benchmark datasets: Labelme \cite{russell2008labelme} and MIRFlickr-25K \cite{huiskes08}. Table~\ref{tab:datasets} demonstrates the detailed statistics. These two datasets, especially MIRFlickr-25K, are rather noisy, with a number of the tags being misspelled or meaningless. Hence, a pre-processing procedure is performed. We match each tag with entries in the Wikipedia thesaurus and only retain the tags in accordance with Wikipedia.

\vspace{-1mm}
\begin{table}[!h]
\centering
    \caption{Statistics of 2 Datasets }     
    \label{tab:datasets}

    \begin{small}
    \begin{tabular}{c|c|c}
    {\bfseries Statistics}     & {\bfseries Labelme} & {\bfseries  MIRFlickr-25K}        \\
    \hline
    No. of images                    &   2,900        &	25,000	                        \\
    \hline
    Vocabulary Size                    & 495            &  1,386	                        \\
    \hline
    Tags per Image (mean/max) 	         & 10.5/48        & 12.7/76                           \\
    \hline
    Images per Tag (mean/max)     &  67.1/379      & 416.5/76,890	                    \\
    \end{tabular}
    \end{small}
\end{table}
\vspace{-1mm}
We compare our method with the state-of-the-art methods, including matrix completion-based models (i.e. LRES \cite{zhu2010image}, TCMR \cite{feng2014image}, RKML \cite{feng2013large}), search-based models (i.e. JEC \cite{makadia2008new}, TagProp \cite{guillaumin2009tagprop}, and TagRelevance \cite{li2009learning}), mixture models (i.e. CMRM \cite{jeon2003automatic} and MBRM \cite{feng2004multiple}) and co-regularized learning model (FastTag \cite{chen2013fast}). The tag refinement procedure by itself, denoted as SMC$\_$IMC, is also compared to verify the benefit from the tag completion procedure.
We tune the parameters
on the validation set of the two datasets separately for every method in comparison.
Note that
the weighting parameter $\mu$ we tune changes from $0.4$ (Labelme) to $0.7$ (MIRFlickr-25K), confirming that as the data become more and more
noisy, we should pay more attention to the noisy tags and less on missing tags.

We measure all the methods in terms of \emph {average precision}$@N$ ($AP@N$) and \emph {average recall}$@N$ ($AR@N$). In the top $N$ completed tags, \emph {precision}$@N$ is to measure the ratio of correct tags and \emph {recall}$@N$ is to measure the ratio of missing ground-truth tags, both averaged over all test images.
\subsection{Evaluation and Observation}
Table~\ref{tab:labelme} and Table~\ref{tab:mirflickr} show performance comparisons on the two datasets, respectively.
\begin{table}[!h]
\centering
    \caption{Performance Comparison on Labelme}     
    \label{tab:labelme}
    \begin{small}
    \begin{tabular}{l|c|c|c|c|c|c}
     & \multicolumn{6}{c}{Labelme}\\
     \cline{2-7}
     & \multicolumn{2}{c|}{N = 2}  & \multicolumn{2}{c|}{N = 5} & \multicolumn{2}{c}{N = 10}\\
     \cline{2-7}
    {\bfseries  } &  {\bfseries  AP}   & {\bfseries AR }  & {\bfseries AP }   & {\bfseries AR }  & {\bfseries AP }   & {\bfseries AR }  \\
    \hline
    \hline
    \bfseries SMC   &  \bfseries 0.51  & \bfseries 0.36   & \bfseries 0.46 &\bfseries  0.50 & \bfseries 0.35 & \bfseries0.62 \\
    \hline
    \bfseries SMC$\_$IMC   &   0.47  &  0.34   & 0.40 & 0.48  & 0.31 & 0.59 \\
    \hline
    \bfseries LRES \cite{zhu2010image}   &   0.42  &  0.32   & 0.35 & 0.45  & 0.27 & 0.56 \\
    \hline
    \bfseries TCMR \cite{feng2014image}   &   0.44  &  0.32   & 0.37 & 0.45  & 0.29 & 0.55 \\
    \hline
    \bfseries RKML \cite{feng2013large}   &   0.21  &  0.14 & 0.19 & 0.20 & 0.14 & 0.22 \\
    \hline
    \cline{1-7}
    \bfseries JEC \cite{makadia2008new}    &  0.33  &  0.29   & 0.27 &0.38  & 0.20 & 0.48 \\
    \hline
    \bfseries TagProp \cite{guillaumin2009tagprop} &  0.39  &  0.31  &0.33 &0.45 &0.25 &0.56 \\
    \hline
    \bfseries TagRel \cite{li2009learning} &   0.43  &   0.32  & 0.34 & 0.45  & 0.27 & 0.55 \\
    \hline
    \cline{1-7}
    \bfseries CMRM \cite{jeon2003automatic}    &  0.20  &  0.14 & 0.18 & 0.19  &0.12 &0.22\\
    \hline
    \bfseries MBRM \cite{feng2004multiple}   &  0.23  &  0.14  & 0.18 &0.20  & 0.12 & 0.27 \\
    \hline
    \cline{1-7}
    \bfseries FastTag \cite{chen2013fast}   &   0.43  &  0.34    & 0.37 & 0.44  & 0.28 & 0.57 \\
    \end{tabular}
    \end{small}
\end{table}

\begin{table}[!h]
\centering
    \caption{Performance Comparison on MIRFlickr-25K}     
    \label{tab:mirflickr}
    \begin{small}
    \begin{tabular}{l|c|c|c|c|c|c}
    & \multicolumn{6}{c}{MIRFlickr-25K}\\
    \cline{2-7}
     & \multicolumn{2}{c|}{N = 2} & \multicolumn{2}{c|}{N = 5} & \multicolumn{2}{c}{N = 10}\\
     \cline{2-7}
    {\bfseries } & {\bfseries  AP}   & {\bfseries AR }  & {\bfseries  AP}   & {\bfseries AR }  & {\bfseries AP }   & {\bfseries AR }  \\
    \hline
    \hline
    \bfseries SMC    &  \bfseries 0.53  & \bfseries 0.39   & \bfseries 0.40 &\bfseries 0.47  &\bfseries 0.33 &\bfseries 0.61 \\
    \hline
    \bfseries SMC$\_$IMC   &   0.45  &  0.34   & 0.36 & 0.43  & 0.30 & 0.52 \\
    \hline
    \bfseries LRES \cite{zhu2010image}   &   0.43  &   0.35  &  0.32 &0.40 & 0.26 & 0.45\\
    \hline
    \bfseries TCMR \cite{feng2014image}   &   0.45  &  0.35   & 0.35 & 0.41 & 0.28 & 0.48 \\
    \hline
    \bfseries RKML \cite{feng2013large}   &   0.21  &  0.15   & 0.13 & 0.23  & 0.13 & 0.22 \\
    \hline
    \cline{1-7}
    \bfseries JEC \cite{makadia2008new}    &  0.33  &  0.30   & 0.25 &0.34  &0.19  & 0.35 \\
    \hline
    \bfseries TagProp \cite{guillaumin2009tagprop}  &  0.39  &   0.35  &0.28 &0.37  &0.20 &0.41 \\
    \hline
    \bfseries TagRel \cite{li2009learning} &  0.42  &  0.34   &0.30 &0.37  &0.20 & 0.40 \\
    \hline
    \cline{1-7}
    \bfseries CMRM \cite{jeon2003automatic}    &  0.20  &  0.15  & 0.13 &0.18 &0.11 & 0.20 \\
    \hline
    \bfseries MBRM \cite{feng2004multiple}   &  0.22  &  0.16   & 0.13 & 0.18  & 0.10 & 0.22 \\
    \hline
    \cline{1-7}
    \bfseries FastTag \cite{chen2013fast}   &   0.43  &  0.35  & 0.30 & 0.41 & 0.27 & 0.42 \\
    \end{tabular}
    \end{small}
\end{table}

We can observe that: $1)$ Generally, methods achieve better performance on Labelme, since tags in MIRFlickr-25K are more noisy. $2)$ Methods based on matrix completion, such as SMC, LRES and TCMR, usually achieve the best performances. $3)$ Our SMC framework shows increasing advantage to LRES as the data become more and more noisy, justifying our assumption and model on the noises. $4)$ SMC nearly outperforms all the other algorithms in all cases. $5)$ Performance comparison between SMC and SMC$\_$IMC demonstrate the remarkable benefit of tag completion for tag refinement. $6)$ Performance on MIRFlickr-25K in some sense provides an evidence for the robustness of SMC.
\section{Conclusion}
In this work we present an effective framework for image annotation by performing tag completion and tag refinement sequentially. Our method first clusters images using sparse subspace clustering and shares tags using a neighbor voting algorithm, then refines tags by adapting inductive matrix completion while novelly utilizing visual and semantic information.
Experiments show the effectiveness of our framework and suggest that tag refinement can benefit a lot from performing tag completion first.

\vspace{-1mm}
\section*{Acknowledgments}

Zhouchen Lin is supported by National Basic Research Program of China (973 Program) (grant no. 2015CB352502),
National Natural Science Foundation (NSF) of China (grant nos. 61272341 and 61231002),
and Microsoft Research Asia Collaborative Research Program.

\vspace{-4mm}
%
\bibliographystyle{abbrv}
\bibliography{sigproc}  
%
%
\end{document}